\documentclass[runningheads]{llncs}
\usepackage{graphicx}
\usepackage{xurl}
\usepackage{cite}
\usepackage[hyperpageref]{backref}
\graphicspath{ {./images/} }
\usepackage{multirow}
\begin{document}
\title{BERT-Based Sentiment Analysis: A Software Engineering Perspective\thanks{All authors have contributed equally.}}
%
%\titlerunning{Abbreviated paper title}
% If the paper title is too long for the running head, you can set
% an abbreviated paper title here
%
\author{Himanshu Batra\and
Narinder Singh Punn\and
Sanjay Kumar Sonbhadra\and
Sonali Agarwal}

\authorrunning{Batra et al.}

\institute{Indian Institute of Information Technology Allahabad, India\\
\email{\{mit2019119,pse2017002,rsi2017502,sonali\}@iiit.ac.in}}

\maketitle              % typeset the header of the contribution
\begin{abstract}
Sentiment analysis can provide a suitable lead for the tools used in software engineering along with the API recommendation systems and relevant libraries to be used. In this context, the existing tools like SentiCR, SentiStrength-SE, etc. exhibited low f1-scores that completely defeats the purpose of deployment of such strategies, thereby there is enough scope for performance improvement. Recent advancements show that transformer based pre-trained models (e.g., BERT, RoBERTa, ALBERT, etc.) have displayed better results in the text classification task. Following this context, the present research explores different BERT-based models to analyze the sentences in GitHub comments, Jira comments, and Stack Overflow posts. The paper presents three different strategies to analyse BERT based model for sentiment analysis, where in the first strategy the BERT based pre-trained models are fine-tuned; in the second strategy an ensemble model is developed from BERT variants, and in the third strategy a compressed model (Distil BERT) is used. The experimental results show that the BERT based ensemble approach and the compressed BERT model attain improvements by 6-12\% over prevailing tools for the F1 measure on all three datasets.
\end{abstract}
\section{Introduction}
%\subsection{A Subsection Sample}

Sentiment analysis is a technique to classify sentences to positive, negative or neutral based upon the point of view expressed through that text. Studies reveal that there are several applications of sentiment analysis for software engineering (SE) like analysing developers' opinions through Stack Overflow posts, the scope of improvement in a library from the comments on GitHub issues section~\cite{article38}, tracking negative comments to inspect bugs in API~\cite{article37}. 

The existing research shows that off-the-shelf sentiment analysis tools~\cite{article22,article27} when tested on SE domain-specific texts tend to disagree with each other. Lin et al.~\cite{article1} reported results of five tools: SentiStrength, NLTK, Stanford CoreNLP, SentiStrength-SE, Stanford CoreNLP SO on app reviews, Stack Overflow and JIRA issues dataset, with the f1-scores ranging from 0.15 to 0.5. SEntiMoji~\cite{article4}, based upon emoji labelled posts, reported an f1-score of 0.47 for the positive sentences and 0.64 for the negative sentences. But still, the utility of the prevailing tools remains quite low for software engineering applications.

The present paper focuses on the scope of improvement of the existing work for the software engineering domain through sentiment analysis. Specifically leveraging the current state-of-the-art performance of BERT (Bidirectional Encoder Representations from Transformers), which has been widely used for major NLP tasks like sentiment analysis, question answering, translation, etc.

\subsection{BERT}
\begin{figure}
\centering
\includegraphics[width=8cm]{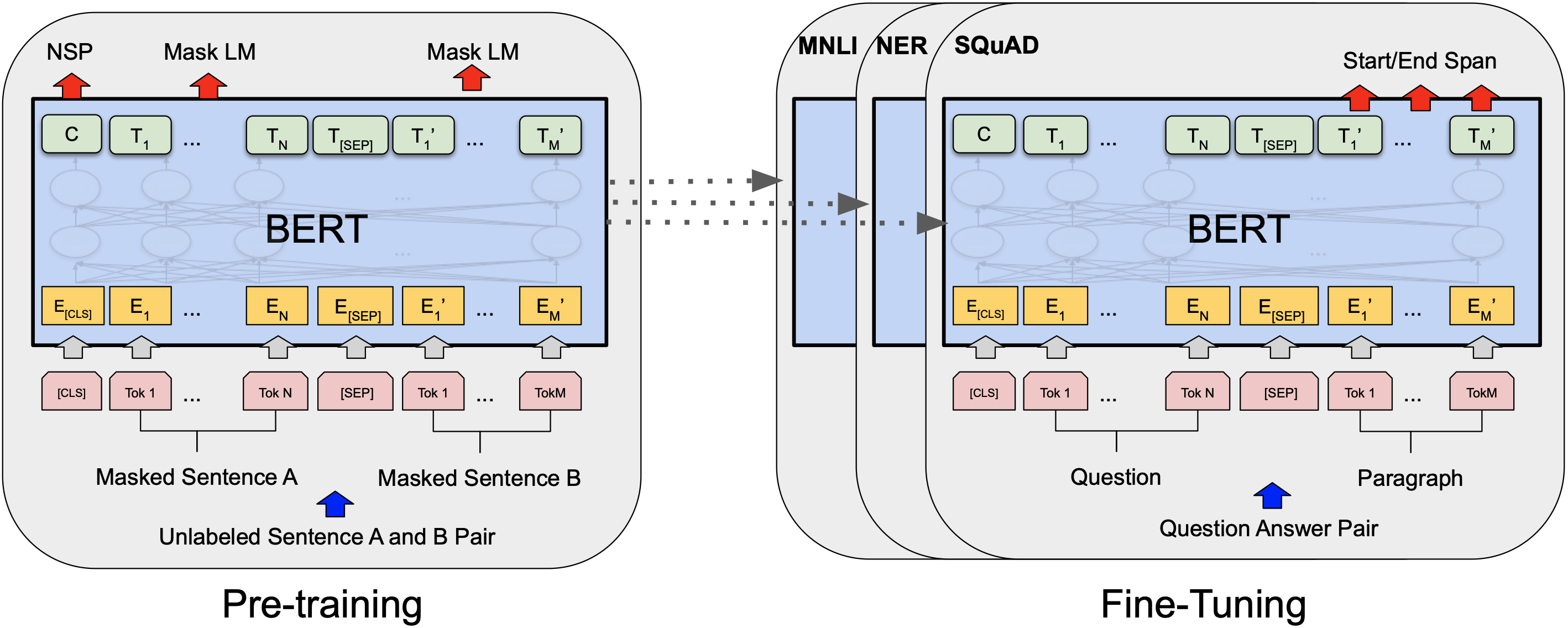}
\caption{Pre-training and fine-tuning procedures for BERT~\cite{article15}.} \label{fig1}
\end{figure}

BERT~\cite{article15} was introduced in 2018 by Devlin et al., for language modelling, based on bidirectional training of transformer (attention model)~\cite{article13}. It uses the attention mechanism that gets the actual context of the text and has two important parts: the encoder (to get the input text) and the decoder (that produces the output or the prediction of the task), through which the pre-training and fine-tuning is carried out as shown in Fig~\ref{fig1}. There are two major techniques to train the BERT model:
\begin{itemize}
    \item \textit{Masked language modelling}: This technique is used to mask 15\% of the words randomly in a sentence and the model tries to predict the masked word based upon the context of the other (non-masked) words in the sentence.
   \item \textit{Next sentence prediction}: In this, the model receives pairs of sentences as input and learns to predict if the second sentence in the pair is the subsequent sentence in the original document. During training, 50\% of the inputs are a pair in which the second sentence is the subsequent sentence in the original document, while in the other 50\% a random sentence from the corpus is chosen as the second sentence.

\end{itemize}

Insufficiency of task-specific labeled data can be handled by BERT through pre-training. This pre-training is done on a large corpus of text ($\sim$2500 million words), which forms a common architecture for a generic task (sentiment analysis, question answering, translation, etc.). This model can then be fine-tuned for our downstream task, to analyse sentiments. On sentiment treebank dataset~\cite{article40} BERT attained a GLUE score of 94.9, considered as a benchmark dataset, it has fine-grained sentiment labels for 215,154 phrases in the parse trees of 11,855 sentences. Following the state-of-the-art potential of BERT model, the present research proposes to exploit the features of BERT variants with the ensemble approach that uses three variants of BERT i.e, the base BERT model, RoBERTa and ALBERT. Furthermore, a compressed BERT model (Distil BERT) is used to establish more robust results for sentiment analysis from the perspective of software engineering.

With the objective and contributions highlighted above, the rest of the paper is organized into several sections. The literature review in section 2 presents the recent developments in the deep learning classification task approaches followed by the methods in section 3 to cover the background knowledge and proposed methodology. The later section 4 covers the exhaustive experimental trials over multiple datasets followed by the results in section 5. Finally, the concluding remarks are presented with future research directions in section 6.

\section{Related work}
With the advent of advancements in the deep learning algorithms, many state-of-the-art frameworks are developed across various domains such as healthcare, natural language processing, target feature learning, etc.~\cite{punn2020multi, torfi2020natural, rajora2021web, punn2020chs}. In consideration of sentiment analysis for software engineering domain,~\cite{article8,article9, article}, mainly focus on deriving developers' opinions/emotions, along with the context. It is observed that the existing tools make dataset driven predictions, where each prediction conflicts with one another~\cite{article11}. Table~\ref{tab1} shows the techniques used for the tools along with the datasets over which the training was performed.
\begin{table}
\caption{Prevailing tools.}\label{tab1}
\centering
 \begin{tabular}{|l|l|l|} 
 \hline
 \textbf{Tool} & \textbf{Dataset used} & \textbf{Technique}  \\  
 \hline
 SentiStrength~\cite{article45} & MySpace informal short texts & Lexicon based  \\ 
 \hline
 SentiStrength-SE~\cite{article44} & Jira comments & Lexicon based  \\
 \hline
 SentiCR~\cite{article2} & Oracle code review comments & Supervised learning \\
 \hline
 Senti4SD~\cite{article3} & Stack Overflow & Supervised learning  \\
 \hline
 NLTK~\cite{article22} & Social media texts & Lexicon based  \\ 
 \hline
 Stanford CoreNLP~\cite{article27} & Movie reviews & Recursive neural tensor network  \\
 \hline
\end{tabular}
\end{table}

The results of NLTK~\cite{article22} and Stanford CoreNLP~\cite{article27} were not very promising as they were not trained on SE-specific datasets and lead to diverging conclusions. SentiCR~\cite{article2} leveraged boosting algorithms and was based on oracle code review comments but only had two classes i.e., negative and non-negative, unlike others classifying on three classes (positive, negative, and neutral). And it provided the highest f1-score among all the datasets. Senti4SD~\cite{article3} being trained on the Stack Overflow dataset with a supervised learning approach had the highest f1-score. Studies have reflected that only to a certain extent these SE-specific tools for sentiment analysis agree with each other. To examine the performance of these tools, Novielli et al.~\cite{article11} performed a benchmark study on standard datasets (Jira and Stack Overflow comments) and to identify potential weak points for improvement and recorded an f1-score of 0.85 on SentiCR and on SentiStrength-SE~\cite{article44} it was 0.83 for the Stack Overflow dataset. 

Along with the ones mentioned in the Table~\ref{tab1}, there are two more related tools, Emotxt~\cite{article25}, based on classifying emotions (anger, sadness, love, joy, fear) and is trained on the Jira and Stack Overflow datasets; whereas, DEVA~\cite{article10} is a dictionary-based approach to detect valence and arousal in the text that can capture individual emotional states (depression, relaxation, excitement, stress, and neutrality). However, in contrast, BERT with task-specific fine-tuning can provide much better results.

Several other convenient techniques help to derive the sentiment from a text. Data augmentation~\cite{article43} is one of such techniques that have been used for various downstream tasks and some of the data augmentation techniques used widely are - Lexical substitution of words in the text, which is basically thesaurus based substitution (replacing a word with its synonym) and word-embedding substitution, this technique primarily takes the pre-trained word embeddings to consideration for the replacement of any word in a text through the nearest neighbor in the embedding space (Word2Vec~\cite{article31}, GloVe~\cite{article32}, FasText~\cite{article41}, Sent2Vec~\cite{pgj2017unsup}). Alternatively, masking of a particular word in a sentence is also a great way to predict a word that has the relative context (BERT). Xie et al~\cite{article29} used a method called back translation for augmentation of unlabelled text on the IMDB dataset for a semi-supervised model which had only 20 labelled examples. Back translation follows translating an English sentence to some other language (say Spanish) and then translating it back to get the new sentence and that sentence is taken as an augmented sentence for training. However, the utility of the prevailing tools remains quite low for software engineering applications. Following this, the present research work exploits the state-of-the-art potential of the BERT variants for sentiment analysis in the software engineering domain.

\section{Proposed work}
 As per the above discussion, it is evident that software engineering sentiment analysis has always been a challenging task. In this context, the present research proposes a robust solution, where initially, the data augmentation is chosen as a pre-processing strategy to aid in training or fine-tuning the BERT variants for efficient results. In the fine-tuned model an additional untrained layer on top of the BERT model is provided for task-specific learning whereas an ensemble approach is proposed that combines the fine-tuned models with a weighted voting scheme. The sentiment analysis is further extended using a compressed model (Distil BERT) to establish robust results.
 
 \subsection{Data augmentation}
 Data augmentation essentially helps to increase the diversity of the dataset that is available for training the model, without actually getting new data or rather scraping data from relevant websites like Stack Overflow, Jira, GitHub, etc. The available techniques used for text data augmentation are as follows:
\subsubsection{Lexical based substitution}
This technique is based on taking into account the synonym of a random word from a given sentence and substituting it using a thesaurus. Similarly, one basic approach is to provide substitution based on word-embeddings, i.e. to replace a word with its nearest neighboring word in the embedding space. 
\subsubsection{Back translation} It is primarily the translation of a sentence to a different language and then translating it back to the original language, where the sentence translated back is likely to use similar words from the language and the context remains unchanged.

 \subsection{Fine-tuning BERT model}
 BERT can be used to get some high-quality language features and can be fine-tuned for the classification task performed on the software engineering datasets to achieve promising results. Fine-tuning provides a few advantages like much faster development, lesser epochs to train on, and limited data requirements. With the pre-trained model, an additional untrained layer of neurons is appended, to fine-tune the model with 2-4 training epochs as suggested in the original BERT paper. Following this three BERT variants are fine-tuned using multiple datasets for better comparative analysis of the proposed strategies for sentiment analysis in software engineering.
 
 \subsection{Ensemble BERT models}
\begin{figure}
\centering
\includegraphics[width=\linewidth]{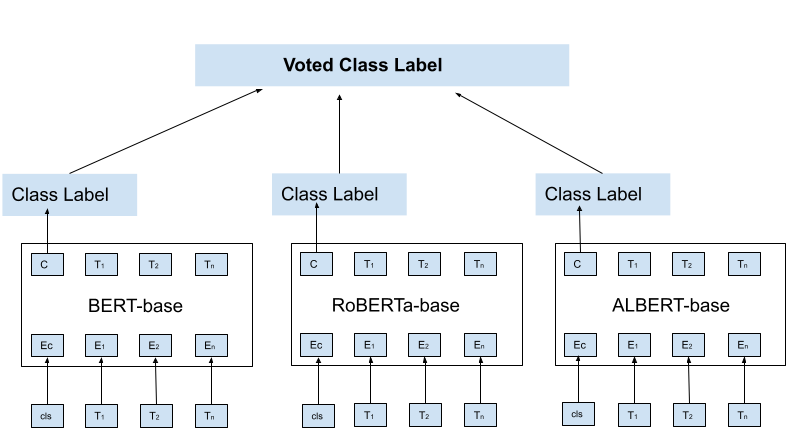}
\caption{BERT model variations used for the ensemble technique } \label{fig2}
\end{figure}

 Based on the fact that combining several models could produce a stronger model hence an ensemble technique is utilized to aggregate the performance of the BERT variants. The scope of prediction is much diverse in the case of the ensemble because there might be certain diversified conclusions on each of the models (that are part of the ensemble), as the core functioning is different for each model. Each model has been pre-trained on a different language modelling task like next sentence prediction used in BERT, sentence order prediction used in ALBERT and dynamic masking used in RoBERTa. Hence it is evident that a stronger model is obtained and the final prediction can be fetched through a voting scheme. For the ensemble technique, BERT base model~\cite{article15}, RoBERTa~\cite{article16}, and ALBERT~\cite{article18} models are used, as shown in Fig~\ref{fig2}. 
 This article used a weighted voting scheme, where the confidence score (aggregated) is considered, i.e. the last Softmax layer output, for each model in the ensemble to get the final weighted prediction. 
 
 \subsection{Compressed model}
 Apart from quantization and model pruning, a distilled model is really helpful to boost up the model's performance rather than to train the model from scratch which requires a large amount of data. The distillation framework mainly contains two models: a larger one and a smaller one. The smaller one mimics the working of the larger one by learning from the probabilities generated by the larger one before the final activation function and produces comparable results. These probabilities are helpful for the generalization capabilities of the compressed model as they are still higher than absolute zero values. Hence, this compressed model is different from the base BERT wherein the model suppresses the lower probability for a particular class and enhances the value for the higher probability class, in its final distribution. The Distil BERT model retains 97\% of the language understanding capabilities and the size is reduced by 40\% in contrast to BERT base model.

\section{Experiments}
\begin{figure}
\centering
\includegraphics[width=\linewidth]{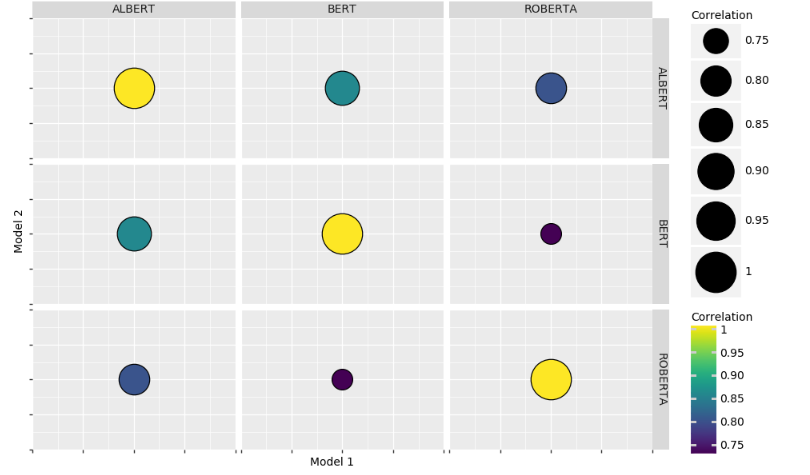}
\caption{Correlation plot for the three BERT variants.} \label{fig3}
\end{figure}

The experiments are aligned to have an overview of the performance on different datasets~\cite{article1,article34} and to check: 1) how the model performs after fine-tuning, 2) how well the BERT models perform when used as an ensemble, and 3) to test the performance of the Distil BERT model. The datasets used for the analysis are GitHub commit comments, Jira issue comments, and Stack Overflow posts. Furthermore, a publicly available synthetic dataset was used to check the performance of the model. For the training part, the number of unique sentences present is 150,000 and the test dataset has 30,000 sentences. It's an added advantage to train and test the model on such a huge dataset, as compared to the Stack Overflow, Jira and GitHub datasets with limited sentences.  
The complete training part is done in a cloud environment having high-end NVidia GPUs. 
Fig~\ref{fig3} shows the correlation among the three BERT models used for ensemble strategy. The correlation scores for RoBERTa-ALBERT and ALBERT-BERT is high, which signifies the models have high probability to generate similar predictions in the ensemble approach.

\begin{table}[]
\centering
\caption{Train, validation and test split for the three datasets.}
\label{tabn}
\begin{tabular}{|l|l|l|l|}
\hline
\textbf{Dataset}        & \textbf{Train data} & \textbf{Validation data} & \textbf{Test data} \\ \hline
Github         & 4480       & 498             & 2137      \\ \hline
Stack Overflow & 2787       & 310             & 1326      \\ \hline
Jira           & 3700       & 410             & 1759      \\ \hline
\end{tabular}
\end{table}

\subsection{Training and testing}
The proposed models are trained and evaluated on the training and test set derived from three datasets as shown in Table~\ref{tabn}. The training data is further split into train and validation set for better learning of the model. The training phase is assisted with the earlystopping technique (halt the training process as soon as the performance stops improving) to avoid the overfitting problem. The trained models are then evaluated on the test set using the standard evaluation metrics, precision, recall and f1-score, for multi-class classification (positive, negative, and neutral). These metrics can be computed using the following equations:

\begin{equation}
Precision \,=\,\frac{TP}{TP+FP}
\label{eq_1}
\end{equation}
\begin{equation}
Recall \,=\,\frac{TP}{TP+FN}
\label{eq_2}
\end{equation}
\begin{equation}
F1\mathrm{-Score} \,=\frac{2*Recall*Precision}{Recall+Precision}
\label{eq_3}
\end{equation}

\section{Results and discussion}
Following from the first proposed strategy involving the fine-tuning of the BERT models, the classification performance concerning sentiment analysis is presented in Table~\ref{tab:my-table}. It is observed that from fine-tuning, the lowest f1-score of 0.67 was recorded for the ALBERT model on Stack Overflow dataset. While the RoBERTa model performed better on GitHub and Jira datasets with the f1-scores as 0.91 and 0.83 respectively. 
The ensemble approach on the other hand was a motivation to obtain a stronger model with slightly varied conclusions from RoBERTa, ALBERT, and BERT-base models, where the f1-score for the GitHub dataset is improved to 0.92, as shown in Table~\ref{tab:tab3}.

\begin{table}[]
\centering
\caption{Fine-tuning results for the three available datasets.}
\label{tab:my-table}
\begin{tabular}{|l|l|l|l|l|l|l|l|l|l|l|}
\hline
\multirow{2}{*}{\textbf{Dataset}}                         & \multirow{2}{*}{\textbf{Model}} & \multicolumn{3}{l|}{\textbf{Neutral}} & \multicolumn{3}{l|}{\textbf{Negative}} & \multicolumn{3}{l|}{\textbf{Positive}} \\ \cline{3-11}
                        &         & \textbf{P}    & \textbf{R}    & \textbf{F1}   & \textbf{P}    & \textbf{R}    & \textbf{F1}   & \textbf{P}    & \textbf{R}    & \textbf{F1}   \\ \hline
\multirow{3}{*}{GitHub} & BERT    & \textbf{0.92} & 0.91 & \textbf{0.91} & 0.89 & \textbf{0.91} & 0.90 & \textbf{0.93} & 0.93 & \textbf{0.93} \\ \cline{2-11} 
                        & ALBERT  & 0.85 & \textbf{0.94} & 0.89 & 0.92 & 0.80 & 0.85 & 0.90 & 0.93 & 0.91  \\ \cline{2-11} 
                        & RoBERTa & \textbf{0.94} & 0.91 & \textbf{0.92} & \textbf{0.90} & 0.91 & \textbf{0.90} & 0.91 & \textbf{0.94} & \textbf{0.93} \\ \hline
\multirow{3}{*}{Stack Overflow} & BERT  & 0.85     & 0.86    &  0.85    & \textbf{0.85}     & \textbf{0.87}     & \textbf{0.86}    & \textbf{0.94}     & \textbf{0.92}     & \textbf{0.93}    \\ \cline{2-11} 
                        & ALBERT  & 0.92 & \textbf{0.94} & 0.93 & 0.62 & 0.62 & 0.62 & 0.71 & 0.34 & 0.48 \\ \cline{2-11} 
                        & RoBERTa & \textbf{0.96} & 0.92 & \textbf{0.94} & 0.78 & 0.82 & 0.80 & 0.57 & 0.76 & 0.65 \\ \hline

\multirow{3}{*}{Jira}   & BERT    & \textbf{0.92} & 0.89 & \textbf{0.91} & 0.85 & 0.68 & 0.76 & 0.76 & \textbf{0.94} & 0.84 \\ \cline{2-11} 
                        & ALBERT  & 0.86 & \textbf{0.94} & 0.90 & \textbf{0.86} & \textbf{0.70} & \textbf{0.78} & 0.78 & \textbf{0.94} & \textbf{0.86} \\ \cline{2-11} 
                        & RoBERTa & \textbf{0.92} & 0.90 & \textbf{0.91} & 0.80 & 0.73 & 0.76 & 0.78 & \textbf{0.88} & 0.83 \\ \hline
\multicolumn{11}{l}{*bold values indicate the highest metric value}
\end{tabular}
\end{table}

\begin{table}[]
\caption{Ensemble model results over different datasets.}\label{tab:tab3}
\centering
\begin{tabular}{|l|l|l|l|l|l|l|l|l|l|}
\hline
\multirow{2}{*}{\textbf{Dataset}}        & \multicolumn{3}{l|}{\textbf{Neutral}} & \multicolumn{3}{l|}{\textbf{Negative}} & \multicolumn{3}{l|}{\textbf{Positive}}

\\ \cline{2-10}
               & \textbf{P}       & \textbf{R}       & \textbf{F1}      & \textbf{P}        & \textbf{R}       & \textbf{F1}      & \textbf{P}        & \textbf{R}       & \textbf{F1}      \\
\hline
GitHub         & 0.92    & 0.93    & 0.92    & 0.92     & 0.90    & 0.91    & \textbf{0.93}     &\textbf{0.93}   & \textbf{0.93}    \\
Stack Overflow &\textbf{0.93}    &\textbf{0.92}    &\textbf{0.92}    & 0.91     & 0.91    & 0.91    & 0.77     & 0.79    & 0.78    \\
Jira           & 0.91    & 0.90    & 0.90    & 0.86     & 0.65    & 0.74    & 0.76     & 0.92    & 0.83  \\ 

\hline
\multicolumn{10}{l}{*bold values indicate the highest metric value}
\end{tabular}
\end{table}

\begin{table}[]
\caption{Compressed model results over different datasets.}\label{tab:tab4}
\centering
\begin{tabular}{|l|l|l|l|l|l|l|l|l|l|}
\hline
\multirow{2}{*}{\textbf{Dataset}}        & \multicolumn{3}{l|}{\textbf{Neutral}} & \multicolumn{3}{l|}{\textbf{Negative}} & \multicolumn{3}{l|}{\textbf{Positive}}

\\ \cline{2-10}
                & \textbf{P}       & \textbf{R}       & \textbf{F1}      & \textbf{P}        & \textbf{R}       & \textbf{F1}      & \textbf{P}        & \textbf{R}       & \textbf{F1}      \\
\hline
GitHub         & 0.93    & 0.92    & 0.92    & 0.91     & 0.92    & 0.91    & \textbf{0.93}     &\textbf{0.93}   & \textbf{0.93}    \\
Stack Overflow & 0.87    & 0.84   & 0.86    & 0.85     & 0.88    & 0.86    &\textbf{0.93}     & \textbf{0.94}    & \textbf{0.94}    \\
Jira           & 0.92    & 0.89    & 0.91    & 0.79     & 0.77    & 0.78    & 0.77     & 0.89    & 0.83  \\ 

\hline
\multicolumn{10}{l}{*bold values indicate the highest metric value}
\end{tabular}
\end{table}

The trend with deep learning models is that the training time is directly proportional to the number of parameters. Therefore it is evident that the compressed model can improve the inference time of the model if deployed as a tool. For the Stack Overflow and the Jira datasets, the f1-score improved to 0.88 and 0.84 respectively, with the Distil BERT model, as shown in Table~\ref{tab:tab4}. Additionally, the f1-score recorded for the synthetic data when tested on the ensemble model was 0.85 which proved to be a major improvement.

\section{Conclusion}
From the perspective of software engineering applications, BERT model shows significant improvement on SE-specific datasets for the classification task when compared to the performance of the prevailing tools. This article proposes that the fine-tuning of the pre-trained model is more effective than taking a large corpus of text and training the model from scratch, while also the ensemble approach can boost the overall performance. The compressed model on the other hand produced similar results and is overall advantageous when deployed as a  tool with limited resources. The extensive trials show that the approaches used outperformed the prevailing tools with an improved f1-score of 0.88 and 0.84 for the Stack Overflow and the Jira datasets respectively. Similarly for the GitHub dataset, the f1-score enhanced for positive and negative classes as 0.93 and 0.91 respectively, with an overall f1-score recorded as 0.92. Furthermore, the study shows the results with synthetic data had an overall f1-score of 0.85 on the ensemble model. The paper highlights the improved results based on the BERT model and can further be extended on domain-specific research, while there is still scope to improve these results as the dataset availability is limited concerning the software domain. Bug prediction~\cite{article42} is another important task that could be studied along with the API recommendation and mining developer's opinions on text/comments.

\section*{Acknowledgment}
	We thank our institute, Indian Institute of Information Technology Allahabad (IIITA), India and Big Data Analytics (BDA) lab for allocating the centralised computing facility and other necessary resources to perform this research. We extend our thanks to our colleagues for their valuable guidance and suggestions.
	
\bibliographystyle{splncs04}
\bibliography{sample.bib}

\end{document}